\documentclass{article} 
\usepackage{iclr2020_conference,times}

\usepackage{amsmath,amsfonts,bm}

\def\eqref#1{equation~\ref{#1}}

\def\1{\bm{1}}

\DeclareMathAlphabet{\mathsfit}{\encodingdefault}{\sfdefault}{m}{sl}
\SetMathAlphabet{\mathsfit}{bold}{\encodingdefault}{\sfdefault}{bx}{n}

\usepackage{subcaption}
\usepackage{hyperref}
\usepackage{url}
\usepackage{graphicx}
\usepackage{tabularx}
\usepackage{amsthm}
\usepackage{booktabs}

\graphicspath{ {./images/} }

\usepackage{xcolor}
\definecolor{Green}{RGB}{50,200,50}
\definecolor{Red}{RGB}{200,50,50}

\title{A Simple Framework for Uncertainty in Contrastive Learning}

\author{Mike Wu$^1$ \& Noah Goodman$^{1,2}$ \\
Department of Computer Science$^1$ and Psychology$^2$\\
Stanford University\\
Stanford, CA 94305, USA \\
\texttt{\{wumike,ngoodman\}@stanford.edu}
}

\iclrfinalcopy 
\begin{document}

\maketitle

\begin{abstract}
Contrastive approaches to representation learning have recently shown great promise.
In contrast to generative approaches, these contrastive models learn a deterministic encoder with no notion of uncertainty or confidence.
In this paper, we introduce a simple approach based on ``contrasting distributions'' that learns to assign uncertainty for pretrained contrastive representations. In particular, we train a deep network from a representation to a distribution in representation space, whose variance can be used as a measure of confidence.
In our experiments, we show that this deep uncertainty model can be used (1) to visually interpret model behavior, (2) to detect new noise in the input to deployed models, (3) to detect anomalies, where we outperform 10 baseline methods across 11 tasks with improvements of up to 14\% absolute, and (4) to classify out-of-distribution examples where our fully unsupervised model is competitive with supervised methods.
\end{abstract}

\section{Introduction}
The success of supervised learning relies heavily on large datasets with semantic annotations. But as the prediction tasks we are interested in become increasingly complex --- such as applications in radiology \citep{irvin2019chexpert}, law \citep{wang2013learning}, and autonomous driving \citep{maurer2016autonomous} --- the expense and difficulty of annotation quickly grows to be unmanageable. As such, learning useful representations without human annotation is an important, long-standing problem. These ``unsupervised'' approaches largely span two categories: generative and discriminative. Generative models seek to capture the data density using ideas from approximate Bayesian inference \citep{hinton2006fast,kingma2013auto,rezende2014stochastic} and game theory \citep{goodfellow2014generative,dumoulin2016adversarially}. However, generating unstructured data (e.g. images or text) poses several technical challenges \citep{zhao2017infovae,arjovsky2017wasserstein}. As such, the representations learned in generative models are often not useable in downstream tasks.
Discriminative approaches forgo generation, training encoders on ``pretext'' prediction tasks where the label is derived from the data itself, such as the color of an image \citep{doersch2017multi,doersch2015unsupervised,zhang2016colorful,noroozi2016unsupervised,wu2018unsupervised,gidaris2018unsupervised}.
A recent family of discriminative approaches, called ``contrastive learning,'' find representations by maximizing the mutual information between transformations of exemplars \citep{hjelm2018learning,bachman2019learning,tian2020makes,chen2020simple,wu2020mutual}. When used in downstream tasks, contrastive methods are quickly approaching the performance of fully supervised analogues \citep{hjelm2018learning,he2020momentum,chen2020improved,misra2020self,chen2020simple,grill2020bootstrap}.

\begin{figure}[h!]
    \centering
    \begin{subfigure}[b]{0.54\textwidth}
        \centering
        \includegraphics[width=\textwidth]{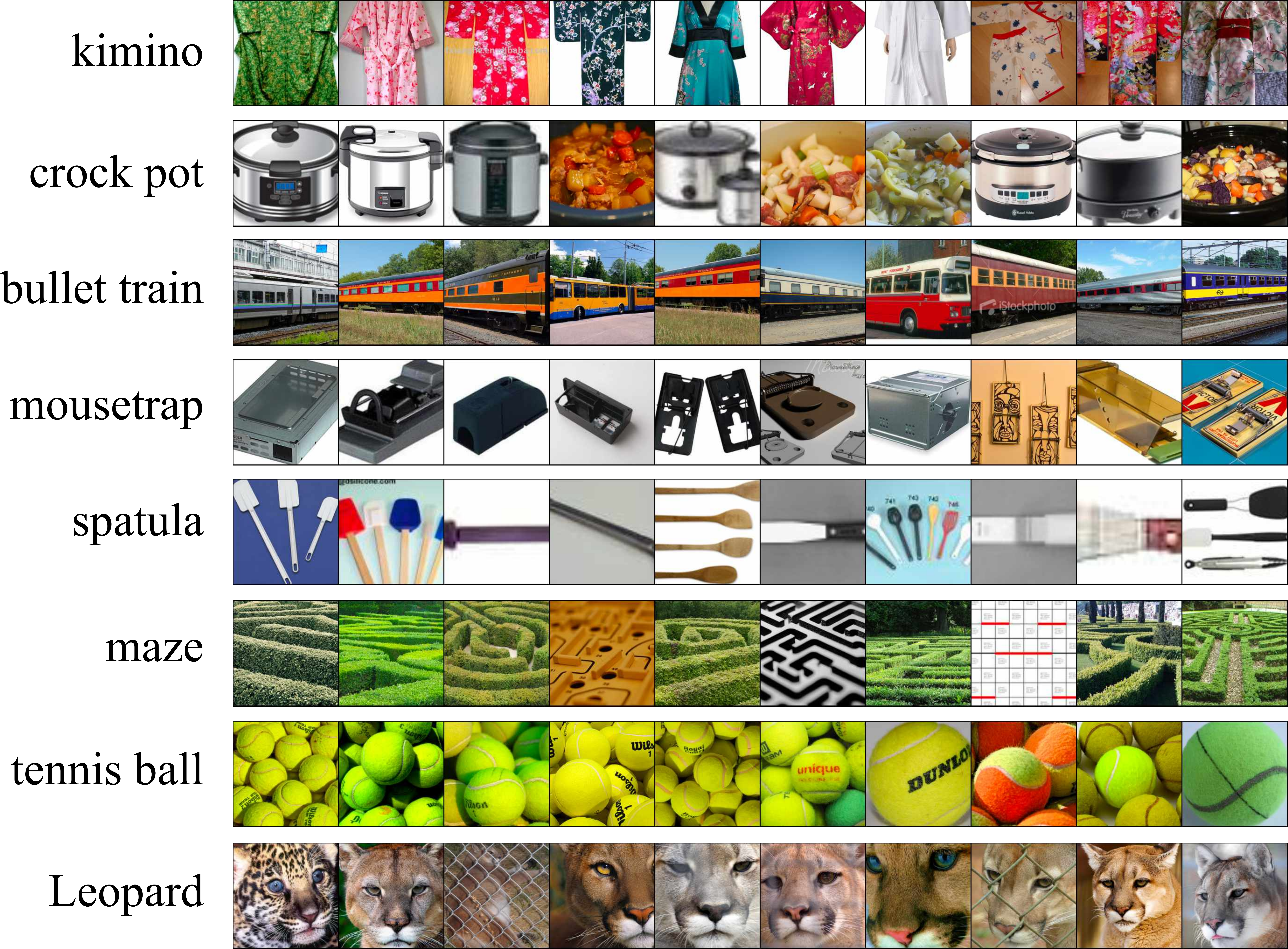}
        \caption{Low DUM Variance Norm}
    \end{subfigure}
    \begin{subfigure}[b]{0.44\textwidth}
        \centering
        \includegraphics[width=\textwidth]{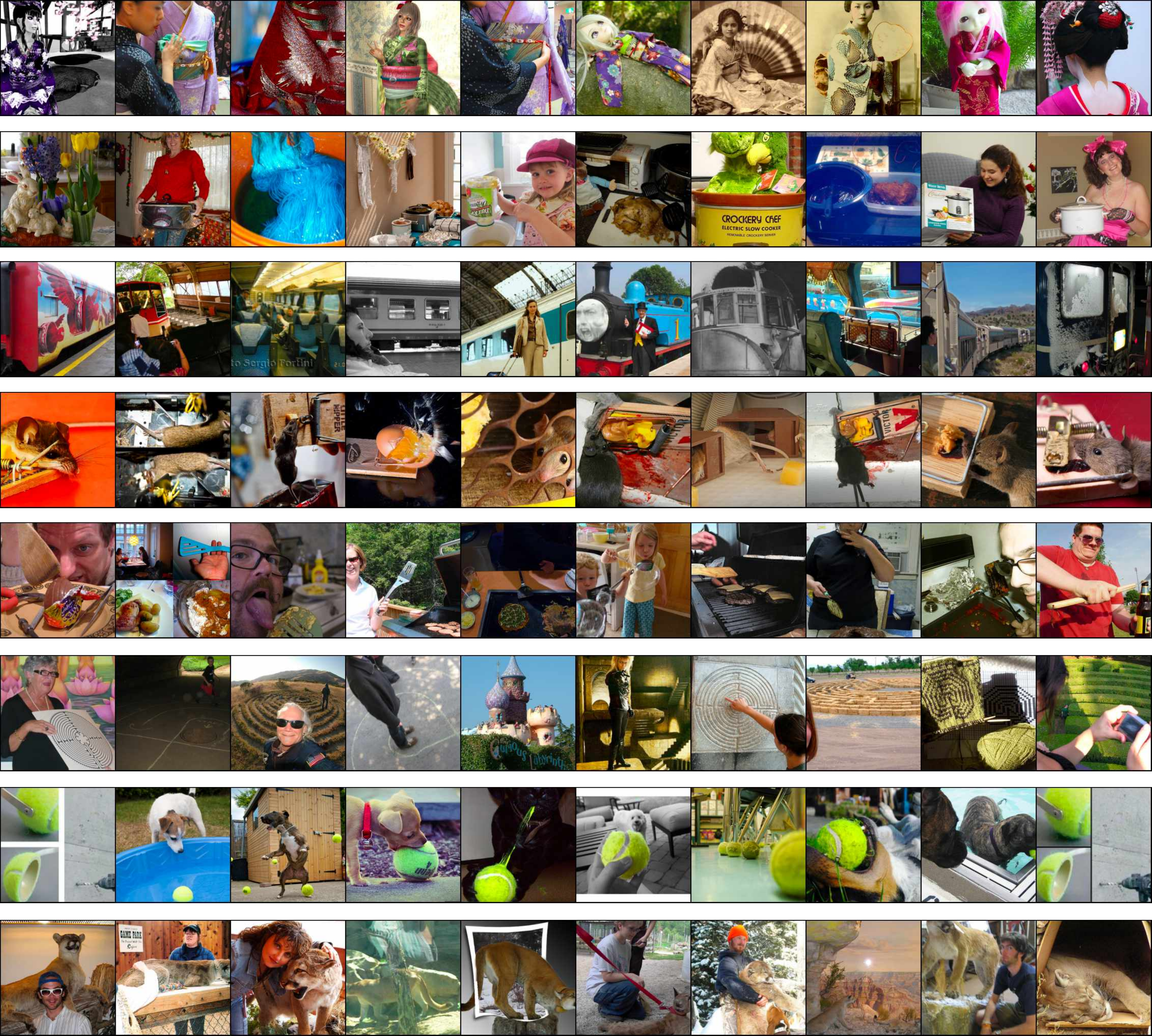}
        \caption{High DUM Variance Norm}
    \end{subfigure}
    \centering
    \begin{subfigure}[b]{0.54\textwidth}
        \centering
        \includegraphics[width=\textwidth]{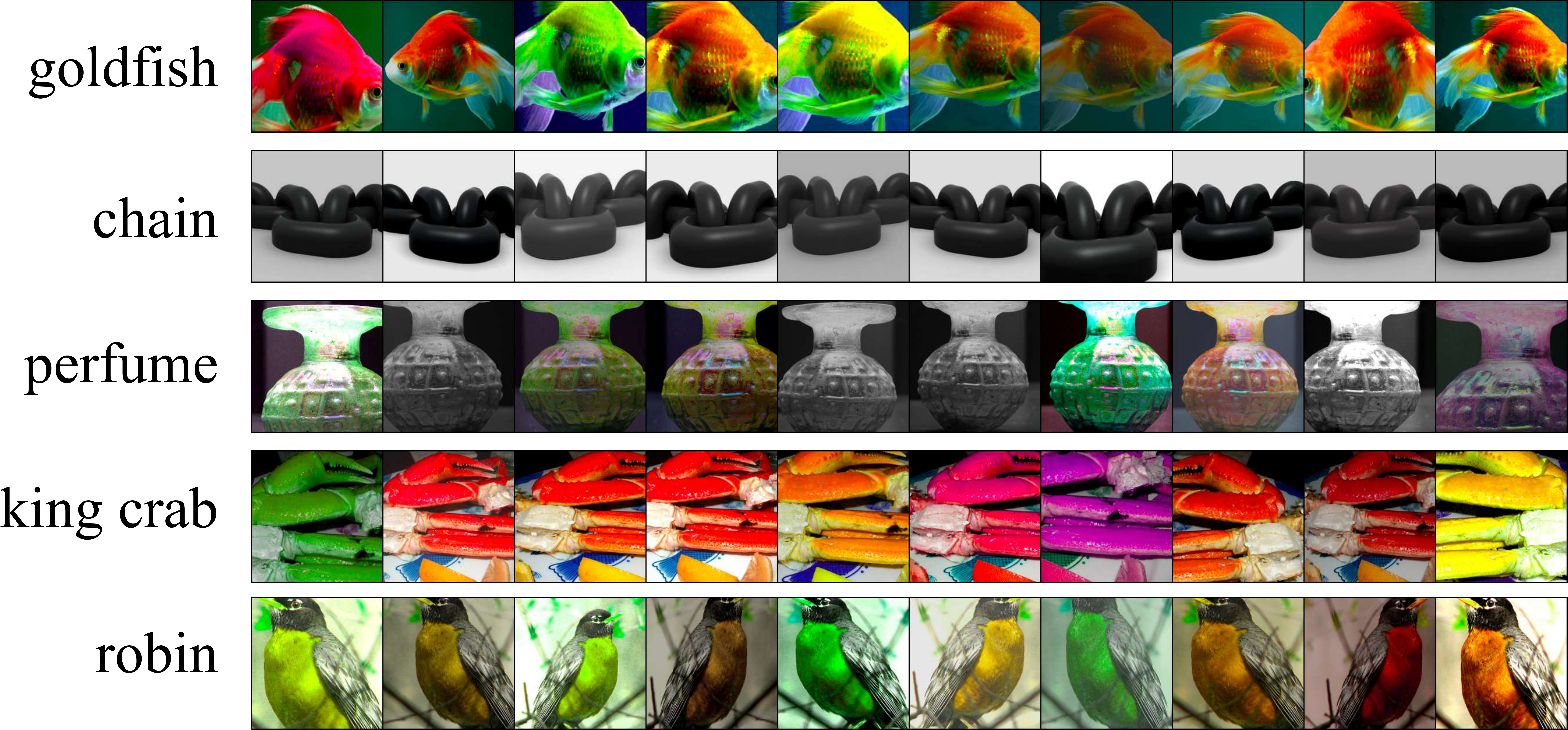}
        \caption{Low DUM Variance Norm}
    \end{subfigure}
    \begin{subfigure}[b]{0.45\textwidth}
        \centering
        \includegraphics[width=\textwidth]{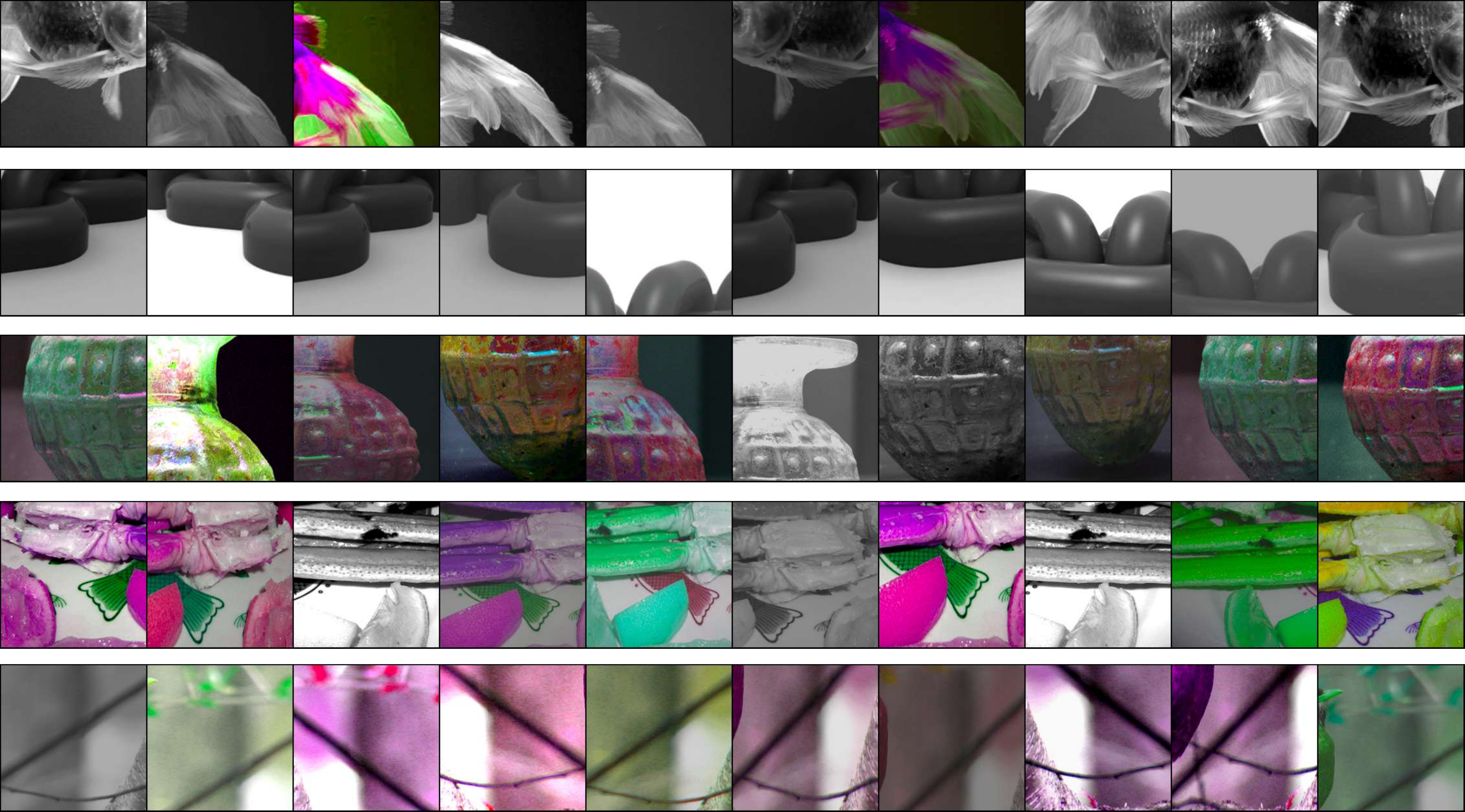}
        \caption{High DUM Variance Norm}
    \end{subfigure}
    \caption{ImageNet images organized by DUM covariance norm. Sub-figures (a) and (b) show images from the same class whereas (c) and (d) show several views of the same image.}
    \label{fig:interp}
    \vspace{-1em}
\end{figure}

Because generative approaches are Bayesian, they provide (though sometimes implicitly) a notion of the posterior uncertainty upon observing and encoding an example.
Contrastive learning, on the other hand, learns a deterministic encoder with no notion of uncertainty.
Such confidence measures can be crucial for detecting domain shift in transfer tasks or in the input to deployed models, and can be extremely useful for practitioners building intuitions about a model and domain.
Our goal in this paper is to learn useful representations of uncertainty on top of pretrained representations.
One intuitive approach would be to treat this as a density estimation problem in the representation space (i.e.~project the dataset under the pretrained encoder and then use standard density estimation models).
However, the objectives of recent contrastive algorithms fundamentally seek embeddings that are uniformly distributed on a compact space \citep{wu2018unsupervised,zhuang2019local,he2020momentum,chen2020simple}.
To the extent that this goal is achieved, density estimation will fail to be useful.
%
In this paper, we formulate an objective based on the task of discriminating the data distribution from other possible distributions.
We use this objective to learn a Deep Uncertainty Model, or DUM, which maps an exemplar embedding to a distribution centered at that embedding.
As such, DUM can be trained on top of the pre-existing embeddings found by popular contrastive frameworks, like SimCLR \citep{chen2020simple}.
We then explore several practical applications of the uncertainties learned by DUM:
\begin{itemize}
    \item \textbf{Interpretability}:  We find the embedding variance to be a good measure of the ``difficulty'' to embed an instance, allowing us to visualize confidence (see Fig.~\ref{fig:interp}).
    \item \textbf{Detecting novel noise}: We show that the variance model can effectively detect corruption of images with novel noise, as might be required in monitoring a deployed classifier.
    \item \textbf{Anomaly detection}: By treating embedding variance as a measure of how unusual an example is, we find improvements up to 14\% absolute over SOTA on 14 datasets.
    \item \textbf{Visual out-of-distribution detection}: We find comparable performance to SOTA supervised methods, despite our OOD algorithm being fully unsupervised.
\end{itemize}

\section{Background}
\label{sec:background}
Suppose we have a dataset $\mathcal{D} = \{x_i\}_{i=1}^n$ of i.i.d samples from $p(x)$, a distribution over a space of natural images $X$.
Let $\mathcal{T}$ be some family of image transformations, $t: X \rightarrow X$, equipped with a distribution $p(t)$.
The common family of transformations (or ``augmentations'') includes random cropping, random color jitter, gaussian blurring, and horizontal flipping \citep{wu2018unsupervised,tian2019contrastive,zhuang2019local,bachman2019learning,he2020momentum,chen2020simple}.

Now, define an encoding function $g_\theta: X \rightarrow \mathbf{R}^d$ that maps an image to an \textit{embedding}, usually parameterized by a deep neural network. The contrastive objective for the $i$-th example is:
\begin{equation}
\mathcal{L}(x_i) = \mathbf{E}_{t,t',t_{1:k} \sim p(t)} \mathbf{E}_{x_{1:k} \sim p(x)} \left[ \log \frac{e^{g_\theta(t(x_i))^Tg_\theta(t'(x_i))/\tau}}{\frac{1}{k+1}\sum_{j\in\{i,1:k\}} e^{g_\theta(t(x_i))^T g_\theta(t_j(x_j))/\tau}} \right]
\label{eq:contrastive}
\end{equation}
where the subscript $x_{1:k} = \{ x_1, \ldots, x_k \}$ and $\tau$ is a scaling hyperparameter.
Intuitively, maximizing Eq.~\ref{eq:contrastive} amounts to pushing the embeddings of two views of the same image close together while pulling the embeddings of two views of different images apart.
If the embedding space is not compact, there are trivial solutions to this objective, hence the typical practice is to L$_2$ normalize the output of the encoder, $g_\theta$.
In this case, the optimal solution is to place the embeddings uniformly across the surface of the unit sphere such that each point is maximally distinctive.

One of the challenges of optimizing Eq.~\ref{eq:contrastive} is that the number of negative samples $k$ is limited by GPU memory, as every negative sample requires additional compute through the encoder. At the same time, a large $k$ is important to improve the stability of Eq.~\ref{eq:contrastive}.
So, different approaches were developed to circumvent this technical hurdle (i.e.~the need to scale up $k$).
One approach, SimCLR \citep{chen2020simple}, cleverly treats the other elements in the same minibatch as negative samples. Since we are already computing the embeddings of all examples in the minibatch, this approach maximally reuses computation.
Although SimCLR relies on using large batch sizes to increase the number of negative examples \citep{chen2020simple}, the benefit over memory-based algorithms \citep{wu2018unsupervised,zhuang2019local,tian2019contrastive,he2020momentum} is that all computation remains on the autodifferentiation tape,
providing more signal to learning a good embedding.
In this work, we will directly build on top of SimCLR to capture uncertainty on instance embeddings. The procedure we propose is very general and can be easily adapted to other contrastive algorithms such as LA \citep{zhuang2019local}, MoCo \citep{he2020momentum}, Deep InfoMax \citep{bachman2019learning}. etc.

\section{The Deep Uncertainty Model}
\label{sec:methods}

The basis for many contrastive learning algorithms were expert-chosen ``pretext tasks'' that were self-supervised by the data itself. In vision, popular pretext tasks included predicting pixel position \citep{doersch2015unsupervised}, image color \cite{zhang2016colorful}, pixel motion \citep{pathak2017learning}, and exemplar identity \citep{dosovitskiy2014discriminative}.
Indeed the contrastive algorithms described in Sec.~\ref{sec:background} arose from the pretext task of determining which instance an augmentation was derived from.

We introduce a pretext task for training a model of \emph{uncertainty} over representations.
The main idea is to \textit{contrast distributions}, rather than contrasting instances.
We start by assuming we have a useful pretrained embedding $g_\theta: X \rightarrow \mathbf{R}^d$, and we represent the push-forward of the data distribution into embedding space by $\hat{p}$ (i.e.~$\hat{p}(z)=p(g_\theta^{-1}(z))$ for an embedding, $z$, of an image $x$).
Let ${\mathcal G}$ be a set of distributions over $\mathbf{R}^d$ with uniformly bounded expectation, that includes $\hat{p}$.
We next define a set, $\Pi \subseteq {\mathcal G}$, of ``negative'' distributions over embeddings that we wish to distinguish $\hat{p}$ from.
Assume that we also have a distribution $p(\pi)$ over the set $\Pi$.
Next, we introduce a \emph{distribution encoder} $f_\phi : {\mathcal G} \rightarrow {\mathcal G}$.
Intuitively we would now like to train $f_\phi$ in order to make the encoding of $\hat{p}$ distinguishable from the encodings of all distributions in $\Pi$.
Inspired by contrastive objectives for instances, the distribution-level contrastive objective for $f_\phi$ is:
\begin{equation}
\mathcal{L}_{\textup{Dist}}(p; \Pi) = \log \frac{e^{\left(\mathbf{E}_{z \sim f_\phi(\hat{p})}[z]\right)^T \left(\mathbf{E}_{z \sim f_\phi(\hat{p})}[z]\right)}}{\mathbf{E}_{\pi \sim p(\pi)} \left[ e^{\left(\mathbf{E}_{z \sim f_\phi(\hat{p})}[z]\right)^T \left(\mathbf{E}_{z' \sim f_\phi(\pi)}[z']\right)}\right]}.
\label{eq:contrast:dist}
\end{equation}
This rather abstract objective does not yet define uncertainty for particular embeddings (and indeed does not yet appear to be useful).
We now make several observations and assumptions that reduce this abstract objective to something more practical.

On the face of it, choosing the set $\Pi$ of negative distributions appears important and difficult.
However, for any choice of $\Pi$, we note that because we assumed uniformly bounded expectations (a mild assumption in the context of neural networks), there exists $b \in \mathbf{R}$ such that:
\begin{align*}
    \mathbf{E}_{\pi \sim p(\pi)} \left[ e^{\left(\mathbf{E}_{z \sim f_\phi(\hat{p})}[z]\right)^T \left(\mathbf{E}_{z' \sim f_\phi(\pi)}[z']\right)}\right] &\leq \sum_{\pi \in \Pi} p(\pi) \cdot e^b = e^b \cdot \sum_{\pi \in \Pi} p(\pi) = e^b
\end{align*}
This in turn implies the bound $\mathcal{L}_{\textup{Dist}}(p; \Pi) \geq \left(\mathbf{E}_{z \sim f_\phi(\hat{p})}[z]\right)^T \left(\mathbf{E}_{z \sim f_\phi(\hat{p})}[z]\right) - b$, suggesting a new objective that does not require specifying $\Pi$ at all.

Our next simplification borrows an idea from meta-learning \citep{choi2019meta,edwards2016towards,hewitt2018variational,oreshkin2018tadam}, treating a bag of i.i.d. samples $z_{1:m} \sim \hat{p}(z)$ as representative of the distribution $\hat{p}$. Thus, $f_\phi$ needs not ingest a distribution but a set of samples instead.
Ideally, the structure of $f_\phi$ should be invariant to the number of representative samples $m$.
We thus decompose $f_\phi$ into two functions: $f^1_\phi$ mapping single samples to distributions and $f^2_\phi$ aggregating these distributions into a single one.

For a single example $x \sim p(x)$, the role of $f^1_\phi$ is to map the embedding $g_\theta(x)$ to a distribution in $\mathcal{G}$.
In practice, we \textit{constrain the distribution returned by $f^1_\phi$ to have mean $g_\theta(x)$.}
Thus, $f^1_\phi$ can only vary the uncertainty. This is absolutely crucial to both preserve the pretrained embeddings and focus the efforts of optimization on capturing uncertainty. Otherwise, embeddings may trivially collapse to a single point. With these changes, the function $f^1_\phi$ finally begins to take on the desired characteristic of representing uncertainty in the embedding of an example.

Now, given $m$ such distributions the role of $f^2_\phi$ is to combine them to a single member of $\mathcal{G}$.
We will assume that $f^2_\phi$ is defined as the product-of-experts distribution \citep{cao2014generalized}. (Loosely motivated by the role that the product of experts (PoE) plays in aggregating the posterior belief distributions from independent observations of a latent variable model \citep{vedantam2017generative,wu2018multimodal}.)
In practice we constrain the output of $f^1_\phi$ to be Gaussian distributions since the product of finitely many Gaussians is itself Gaussian. Precisely, given $m$ Gaussian distributions with corresponding means $\mu_i$ and covariances $\Sigma_i$, the product has mean $\mu_{1:m} = \left(\sum_{i=1}^m \mu_{i} \omega_{i} \right)\left(\sum_{i=1}^m \omega_{i} \right)^{-1}$ and covariance $\Sigma_{{1:m}} = \left(\sum_{i=1}^m \omega_{i} \right)^{-1}$
where $\omega_{i} = (\Sigma_{i})^{-1}$ is the inverse of the $i$-th covariance matrix.
In practice, we use diagonal covariances to simplify inversion.

Now, we reach our final objective on sets of data points:
\begin{equation*}
    \mathcal{L}_{\text{DUM}}(x_{1:2m}) = \mathbf{E}_{t_{1:2m}\sim p(t)}\left[ \left(\mathbf{E}_{z \sim f^2_\phi(\{f^1_\phi(g_\theta(t_i(x_i)))\}_{i=1}^m)}[z]\right)^T \left(\mathbf{E}_{z \sim f^2_\phi(\{f^1_\phi(g_\theta(t_j(x_j)))\}_{j=m}^{2m})}[z]\right) \right]
\end{equation*}
By the bound above, we know $\mathcal{L}_{\text{DUM}}(x_{1:2m}) -b \leq \mathcal{L}_{\textup{Dist}}(p; \Pi)$ for $x_{1:2m} \sim p(x)$ i.i.d.~and any family of distributions, $\Pi$.
As $b$ is a constant, we can ignore it in optimization. As a result, maximizing $\mathcal{L}_{\text{DUM}}$ also maximizes $\mathcal{L}_{\text{Dist}}$.
The expression $f^2_\phi(\{f^1_\phi(g_\theta(t_i(x_i)))\}_{i=1}^m)$ composes all the pieces of DUM described in the last several paragraphs: a set of $m$ representations (via $g_\theta$) of augmented examples $x_{1:m}$ are each given to $f^1_\phi$ to define $m$ Gaussian distributions, which are further combined via $f^2_\phi$ (via PoE) to a single Gaussian distribution, from which we can sample.
We call this simple model with $\mathcal{L}_{\text{DUM}}$ as its objective, a \textit{Deep Uncertainty Model}, or DUM.
Overall DUM is a relatively simple procedure for learning to associate uncertainty to a pretrained representation model.
Like any pretext task, its value will be determined by its downstream use.

\textbf{On the number of representative samples.} We make a final observation on the importance of choosing $m {>} 1$ when using DUM. Suppose instead that $m{=}1$, and sample $x,x' \sim p(x)$. Then, the DUM objective is simply $\mathbf{E}_{t,t'\sim p(t)}\left[\left(\mathbf{E}_{z \sim f^1_\phi(g_\theta(t(x))}[z]\right)^T\left(\mathbf{E}_{z' \sim f^1_\phi(g_\theta(t'(x'))}[z'] \right)\right]$. That is, it is the dot product of the means of the two Gaussian distributions specified by $f^1_\phi$. (Note that $f^2_\phi$ is the identity when $m=1$.) This is problematic as the variance is no longer in the objective. Further, given that we have defined $f^1_\phi$ to center its distribution at the instance representation, optimizing DUM with $m=1$ is truly trivial. Only with $m{>}1$  do we allow variances, and optimizable parameters, to enter the optimization through the PoE.

\section{Applications of Uncertainty}

Having introduced DUM, we focus on three applications of probabilistic embeddings: embedding interpretability, novel noise detection, anomaly detection, and out-of-distribution classification.

\subsection{Embedding Interpretability}
While contrastive methods assume uniformity among examples, we intuitively know that some instances must be more interpretable, and useful, than others. For instance, low resolution images cluttered with blurry objects must be more difficult than a  sharp image with one centered object. As a practitioner, one may wish to have a metric of how (un)certain an embedding is. In more safety-critical domains such as self-driving cars or healthcare \citep{chen2017machine}, the ability to measure model uncertainty is of utmost importance.

We propose using the norm of the DUM covariance matrix (i.e. the variance of the distribution returned by $f^1_\phi(g_\theta(x))$) as a metric of how certain we are of an instance embedding $g_\theta(x)$: the higher the norm, the less certain we are. Using pretrained SimCLR ResNet50 embeddings\footnote{We use SimCLR embeddings from \url{https://github.com/google-research/simclr}.} fit on ImageNet, we train a DUM model to map embeddings to these covariance matricies. Fig.~\ref{fig:interp} shows the examples in the ImageNet dataset with the lowest norms (most certain embeddings) and highest norms (least certain embeddings)  for five randomly chosen classes.

Notably, we observe that images with the most certain embeddings are ``prototypical'' in the object class, often centered and the primary focal point of the image. On the other hand, images with the least certain embeddings are crowded with auxiliary objects, with the primary object often being occluded in the scene. For example, where spatulas in Fig.~\ref{fig:interp}a are displayed in the forefront with monochrome backgrounds, spatulas in Fig.~\ref{fig:interp}b contain food, people, and a variety of backgrounds and lightning. (And recall this is an unsupervised method: it does not know that spatulas are the target.) In the appendix, we show a larger range of classes and show that we can do the same visualizations on transfer distributions, getting similarly appealing results.

Along the same intuition, we do not expect the embeddings of different augmentations (or ``views'') of the same image to be equal in quality. For example, a tiny crop of an image with noise should be much harder to usefully embed than merely a horizontal flip alone.
In Fig.~\ref{fig:interp}c,d, we confirm this hypothesis by visualizing the most certain and least certain image views by DUM variance norm. We note that the most certain views \textit{make little use of crops}, whereas the least certain views are all small crops focusing on local subsets of the image, making it  difficult for even the human eye to recognize the contents of the original image. These findings agree with prior work \citep{tian2020makes,wu2020mutual} which find cropping to be the most difficult (but important) augmentation.

\begin{table}[h!]
\tiny
\centering
\caption{\textbf{Suite of Visual Corruption Detection Tasks.} We compare DUM variance norms for held-out uncorrupted data to data corrupted with a variety of noise. `None' is the baseline of a second held-out set of uncorrupted data.}
\begin{subtable}[h]{0.32\textwidth}
    \centering
    \begin{tabular}{lccc}
    \toprule
    Corruption & P-value & AUROC\\
    \midrule
    None & 0.53 & 50.2 \\
    Gaussian Noise & 0.0 & 71.0 \\
    Shot Noise & 0.0 & 67.2 \\
    Impulse Noise & 0.01 & 57.3\\
    Defocus Blur & 6e-267 & 70.3 \\
    Glass Blur & 0.0 & 70.2 \\
    Motion Blur & 1e-6 & 59.8\\
    Zoom Blur & 0.0 & 66.8 \\
    Snow & 0.0 & 64.2 \\
    Frost & 0.44 & 52.4 \\
    Fog & 0.0 & 64.2 \\
    Brightness & 5e-15 & 63.6 \\
    Contrast & 6e-24 & 63.6 \\
    Elastic & 2e-125 & 66.4 \\
    Pixelate & 6e-34 & 63.7 \\
    JPEG & 0.0 & 67.2 \\
    \bottomrule
    \end{tabular}
    \caption{CIFAR10-C}
\end{subtable}
\begin{subtable}[h]{0.32\textwidth}
    \centering
    \begin{tabular}{lcc}
    \toprule
    Corruption & P-value & AUROC \\
    \midrule
    None & 0.56 & 50.6 \\
    Gaussian Noise & 0.0 & 76.3 \\
    Shot Noise & 0.0 & 71.6 \\
    Impulse Noise & 0.0 & 72.1 \\
    Defocus Blur & 0.0 & 66.1 \\
    Glass Blur & 0.0 & 80.1 \\
    Motion Blur & 2e-158 & 62.2 \\
    Zoom Blur & 0.0 & 68.0 \\
    Snow & 1e-158 & 63.9 \\
    Frost & 0.03 & 54.6 \\
    Fog & 1e-203 & 69.8 \\
    Brightness & 8e-8 & 62.9\\
    Contrast & 3e-8 & 64.6 \\
    Elastic & 1e-16 & 61.6\\
    Pixelate & 2e-5 & 62.8\\
    JPEG & 0.0 & 65.9 \\
    \bottomrule
    \end{tabular}
    \caption{CIFAR100-C}
\end{subtable}
\begin{subtable}[h]{0.32\textwidth}
    \centering
    \begin{tabular}{lcc}
    \toprule
    Corruption & P-value & AUROC \\
    \midrule
    None & 0.34 & 50.7 \\
    Gaussian Noise & 0.0 & 88.3 \\
    Shot Noise & 0.0 & 86.4 \\
    Impulse Noise & 0.0 & 87.2 \\
    Defocus Blur & 0.0 & 81.4 \\
    Glass Blur & 0.0 & 86.9 \\
    Motion Blur & 0.0 & 82.7 \\
    Zoom Blur & 0.0 & 84.4 \\
    Snow & 0.0 & 80.3 \\
    Frost & 0.0 & 77.6 \\
    Fog & 0.0 & 77.1 \\
    Brightness & 0.0 & 66.4 \\
    Contrast & 0.0 & 71.2 \\
    Elastic & 0.0 & 82.0 \\
    Pixelate & 1e-74 & 65.7  \\
    JPEG & 1e-61 & 65.7 \\
    \bottomrule
    \end{tabular}
    \caption{TinyImageNet-C}
\end{subtable}
\label{table:corruptions}
\end{table}
In summary, the DUM covariance norm acts as a statistic for how certain a contrastive model is in its embedding. As a practitioner, for a new instance, we can now gauge the quality of its embedding by comparing its DUM statistic to those of a known set. For instance, human-in-the-loop monitoring of a trained-and-deployed ML system may require a method to alert the user when new sources of noise may have corrupted the incoming data. 
As a demonstration, we consider visual corruptions on image corpora \citep{hendrycks2019benchmarking}, which are used to measure the robustness of classifiers for safety-critical applications. We train SimCLR on the un-corrupted corpora (for TinyImageNet, we train SimCLR on all of ImageNet), on top of which we fit DUM. Here, we explore using the DUM variance norm as a measure of ``typicality'' of a new unseen instance. For a wide class of corruptions, we perform a two sample $t$-test between norms of a held-out set from the training distribution and norms of a corrupted set. We find highly significant differences (Table~\ref{table:corruptions}), indicating that the DUM score is very good at distinguishing the training distribution from noise-corrupted versions.
Moreover, we compute the area under the ROC curve (AUROC) exploring precision vs recall in detecting noisy examples as we vary a threshold on DUM score (examples from the corrupted set are positive labels). Across datasets and noise types we find AUROCs between 60 and 90, indicating that the DUM score can be used to identify corrupted individual examples as well.

\subsection{Anomaly Detection}
\label{section:anomaly}
Not unrelated, we next explore anomaly detection, or identifying instances in a dataset that deviate significantly from the majority of examples \citep{chalapathy2019deep}.
Anomaly detection is an inherently  unsupervised task, in which an algorithm is provided all examples (i.e.~there is no separate test set) and must assign an ``oddity'' score to each.
Furthermore, we seek a domain agnostic method that uses minimal prior information specific to a given dataset.
Here, we propose to characterize anomalies as examples with high uncertainty, as measured by DUM covariance norms.

We consider 14 real-world datasets from the UCI repository \citep{asuncion2007uci} that contain a wide range of sizes and domains. None are image datasets, each containomg vectorized features relevant to their domain e.g. IMU measurements for PAMAP2.
These 14 datasets have been used intensively in anomaly detection literature \citep{sugiyama2013rapid,pham2012near,pang2018learning}. They include a variety of real-world settings, such as the intrusion-detection challenge, \textsc{KDD1999}, one of the most popular benchmarks in the field.
We standardize, by feature, each dataset such that feature values are within the range 0 to 1.
Because we are working with features vectors already, we fit DUM directly on the inputs themselves, rather than learning a SimCLR embedding first. We found comparable performance (see Sec.~\ref{sec:app:simclr}) when using SimCLR embeddings so we prefer the former as it is computationally cheaper.
The DUM encoder is a 3-Layer MLP with 4096 hidden nodes and ReLU non-linearities. We optimize the DUM objective for 100 epochs using Adam \citep{kingma2014adam} with batch size 256 and a learning rate of 1e-3.

\begin{table}[h!]
\centering
\tiny
\caption{\textbf{Suite of Anomaly Detection Tasks.} We compare DUM to 10 baselines on 14 datasets. The symbol ``t/o'' represents a ``timeout'' where model fitting would be too costly for practical use.}
\begin{tabular}{l|l|l|c|c|c|c|c|c|c|c|c|c|c}
\toprule
\multicolumn{14}{c}{Area under the Receiver Operating Characteristic (AUROC)} \\
\midrule
& \# data & \# out & ISO & LOF & SVM & SP & EE & KNN & ABOD & AE & LeSiNN & REPEN & DUM \\
\midrule
Arrhythmia & 452 & 207 & $72.1$ & $73.2$ & $73.6$ & $70.4$ & $69.8$ & $73.4$ & $72.3$ & $73.3$ & $74.0$ & $74.9$ & $\mathbf{76.6}$ \\
CoverType & 286K & 2.8K & $94.9$ & $53.7$ & $93.7$ & $91.3$ & $76.3$ & $76.8$ & $73.7$ & $93.9$ & $93.8$ & $\mathbf{95.1}$ & $91.4$ \\
Ionosphere & 351 & 126 & $86.2$ & $89.4$ & $85.1$ & $88.1$ & $\mathbf{95.1}$ & $93.3$ & $91.7$ & $81.2$ & $89.5$ & $93.1$ & $81.0$ \\
Isolet & 960 & 240 & $84.6$ & $51.7$ & $71.4$ & $52.8$ & $75.1$ & $93.3$ & $91.7$ & $81.1$ & $84.4$ & $93.1$ & $\mathbf{100.0}$ \\
Kdd1999 & 4.8M & 703K & $74.1$ & t/o & t/o & $84.3$ & $69.2$ & t/o & t/o & $72.7$ & $82.5$ & $85.5$ & $\mathbf{98.0}$ \\
MFeat & 600 & 200 & $83.7$ & $54.2$ & $65.7$ & $89.2$ & $52.4$ & $81.9$ & $58.9$ & $90.1$ & $92.4$ & $89.3$ & $\mathbf{99.9}$ \\
OptDigits & 1.7K & 554 & $61.9$ & $59.8$ & $60.8$ & $74.1$ & $77.8$ & $79.1$ & $64.1$ & $59.7$ & $86.9$ & $86.0$ & $\mathbf{99.6}$ \\
PAMAP2 & 373K & 126K & $88.4$ & t/o & t/o & $\mathbf{89.0}$ & $82.6$ & t/o & t/o & $87.6$ & $79.9$ & $88.8$ & $87.5$ \\
PIMA & 768 & 268 & $67.6$ & $60.1$ & $62.4$ & $64.0$ & $67.9$ & $70.8$ & $66.7$ & $63.0$ & $71.7$ & $71.7$ & $\mathbf{81.5}$ \\
Record & 5.7M & 21K & $\mathbf{99.9}$ & t/o & t/o & $\mathbf{99.9}$ & $65.5$ & t/o & t/o & $99.7$ & $99.7$ & $99.7$ & $\mathbf{99.9}$ \\
Skin & 245K & 51K & $62.4$ & $55.5$ & $54.8$ & $65.1$ & $89.2$ & $57.0$ & $52.4$ & $59.3$ & $72.6$ & $75.1$ & $\mathbf{96.2}$ \\
Spambase & 4.6K & 1.8K & $60.7$ & $54.3$ & $53.7$ & $54.7$ & $54.8$ & $50.6$ & $61.4$ & $54.8$ & $57.3$ & $59.0$ & $\mathbf{83.4}$ \\
Statlog & 6.4K & 626 & $74.8$ & $50.2$ & $75.9$ & $54.5$ & $61.1$ & $56.1$ & $50.8$ & $84.5$ & $58.3$ & $68.8$ & $\mathbf{89.2}$ \\
Wdbc & 569 & 212 & $82.5$ & $53.7$ & $69.8$ & $72.9$ & $88.9$ & $77.4$ & $71.3$ & $64.6$ & $82.5$ & $87.9$ & $\mathbf{96.9}$ \\
\bottomrule
\end{tabular}
\label{table:anomaly}
\end{table}

We evaluate our method against 10 baselines, seven of which are classic anomaly detection algorithms: isolation forest or ISO \citep{liu2008isolation}, local outlier factor or LOF \citep{breunig2000lof}, one-class support vector machines or SVM, rapid sampling or SP \citep{sugiyama2013rapid}, elliptic envelope or EE \citep{rousseeuw1999fast}, K-nearest neighbors or KNN, and angle-based outlier detection or ABOD \citep{kriegel2008angle}. As these are well-documented algorithms, we refer to existing literature \citep{sugiyama2013rapid} for an overview.
Additionally, we compare our method to: (1) an autoencoder or AE \citep{kramer1991nonlinear}, which assigns scores based on reconstruction error, (2) Least similar neighbors or LeSiNN \citep{pang2015lesinn}, which classifies anomalies based on distance to neighbor points, and (3) a representation learning algorithm, REPEN \citep{pang2018learning}, which finetunes LeSiNN classes using a triplet loss.

Table~\ref{table:anomaly} reports AUROCs. We find that our method outperforms all 10 baseline algorithms in 11 of the 14 datasets, often by a large margin. For example, our method outperforms the next best model by 14 points on Statlog, 13 points in OptDigits, and 12 points in KDD1999, additionally reaching near perfect AUROC in several datasets. The KDD1999 dataset was built for detecting network intrusions, a difficult problem that requires differentiating between ``good'' connections made by users from ``bad'' connections made by unauthorized attackers. We highlight that an increase from 0.86 to 0.98 AUROC can potentially be the difference between requiring human supervision and allowing a fully automated system.
For the three datasets where our models does not return the best performance, there is no clear pattern to the best performing algorithms -- each dataset is dominated by an algorithm that only performs well on a subset of the other datasets.


\subsection{Out-of-Distribution Image Detection}

Third, we classify out-of-distribution (OOD) images in visual corpora, a task related to but distinct from anomaly detection. In anomaly detection, we assumed that the dataset contains a small contamination of anomalies and the goal is to discriminate between outliers and inliers. Critically, algorithms observe both before making a prediction, despite not knowing which is which.
In visual OOD, we are not given the outliers ahead of time.
Instead, the training data are drawn from some distribution, and the resulting model is used to score a test dataset consisting of unseen examples from both the training distribution and an ``outlier'' distribution.
The quality of an algorithm is in its ability to correctly separate unseen new-domain from unseen old-domain examples.

\begin{table}[h!]
    \centering
    \tiny
    \caption{\textbf{Suite of OOD Image Detection Tasks}. A comparison of our unsupervised approach using DUM covariance norms to six state-of-the-art supervised methods.}
    \begin{tabular}{c|c|c|c|c|c|c|c|c|c}
        \toprule
        \multicolumn{10}{c}{Area under the Receiver Operating Characteristic (AUROC)} \\
        \midrule
        In-dist & Out-of-dist & Baseline & ODIN & Mahalanobis & Res-Flow & Gram & Rotation Pred. & DUM & DUM (Circ)\\
        \midrule
        CIFAR-10 & SVHN & $89.8$ & $96.6$ & $99.2$ & $98.9$ & $99.5$ & $93.8$ & $98.5$ & $83.5$ \\
                 & TinyImageNet & $90.9$ & $93.8$ & $99.5$ & $99.6$ & $99.7$ & $83.8$ & $99.3$ & $84.2$ \\
                 & LSUN & $91.0$ & $94.0$ & $99.7$ & $99.8$ & $99.9$ & $79.3$ & $99.8$ & $84.9$\\
        \midrule
        CIFAR-100 & SVHN & $79.3$ & $93.7$ & $98.4$ & $97.8$ & $96.0$ & $82.1$ & $94.5$ & $53.6$\\
                  & TinyImageNet & $77.0$ & $87.1$ & $98.2$ & $98.9$ & $98.9$ & $66.7$ & $92.5$ & $78.1$ \\
                  & LSUN & $75.5$ & $84.7$ & $98.2$ & $99.1$ & $99.2$ & $73.5$ & $96.8$ & $79.3$ \\
        \midrule
        SVHN & CIFAR-10 & $92.9$ & $92.9$ & $99.3$ & $99.6$ & $97.3$ & $97.3$ & $86.4$ & $72.4$ \\
             & TinyImageNet & $93.5$ & $93.5$ & $99.8$ & $99.9$ & $99.7$ & $97.1$ & $96.1$ & $83.3$ \\
             & LSUN & $91.5$ & $91.5$ & $99.8$ & $100.0$ & $99.8$ & $96.5$ & $97.4$ & $84.3$ \\
        \bottomrule
    \end{tabular}
    \label{table:ood}
\end{table}
Following prior work \citep{hendrycks2016baseline,liang2017enhancing,lee2018simple,zisselman2020deep,hendrycks2018deep,sastry2019detecting,hendrycks2019using}, we consider three datasets --- CIFAR10, CIFAR100 \citep{krizhevsky2009learning}, and SVHN \citep{netzer2011reading} --- whose training split is used to fit a neural network and whose test split acts as the unseen old-domain set.
Then, we explore four OOD datasets: CIFAR10, Tiny ImageNet \citep{russakovsky2015imagenet}, SVHN, and LSUN \citep{yu2015lsun}.
We compare our method against six baselines, \textit{all} of which compute scores derived from pretrained \textit{supervised} networks (i.e. ResNet34) on the training corpus. We emphasize this last distinction: the baseline algorithms rely heavily on knowledge of the known classes in the training distribution and effectively classify an outlier as an instance that does not fit well into any of the these classes. However, we note that assuming such strong supervision is not always tractable as practical applications may not have the resources or knowledge for annotation. The DUM approach is \textit{not} provided with any  knowledge about classes within the training domain.

Table~\ref{table:ood} reports AUROCs. While we usually use diagonal covariance, we also include a version of DUM with circular covariance, denoted by DUM (Circ).
We find that consistently, DUM is competitive with supervised OOD algorithms, despite not having any class information. Further, DUM commonly outperforms Rotation Prediction, the most similar baseline that also relies on a contrastive backbone, albeit supervised (see Sec.~\ref{sec:relwork}). Finally, we also find significant improvements of DUM over DUM (Circ), suggesting the separate dimensions of variance capture useful information.
With these results, we establish a strong baseline for future work in \textit{un}-supervised OOD detection.

\section{Related Work}
\label{sec:relwork}

\textbf{Variational Autoencoders}
The design of DUM is reminiscent of the inference network in variational autoencoders, or VAE \citep{kingma2013auto}. In this setting, given latent variables $z$ and observed variables $x$, we define a family of distributions $\mathcal{Q}$ from which we pick a member $q_{\psi}$ to minimize the evidence lower bound, or the distance between the variational and true posteriors. If we ``amortize'' \citep{gershman2014amortized} the variational posterior across inference queries with different inputs $x$, then we instead write $q_{\psi(x)}(z)$ where we choose the parameters $\psi(x)$ based on the value of $x$. This is exactly the form of $f^1_\phi$ in DUM where $g_\theta(x)$ can be considered a latent variable. VAEs regularize the posterior to be close to a prior distribution whereas DUM is unregularized and greedily finds the distribution to minimize the DUM loss, which is quite different than the VAE loss.

\textbf{Interpretable Neural Networks.}
There has been an extensive body of work of interpreting supervised models. In particular, visualizing examples is a popular method used to qualitatively gauge a model's underlying logic. For instance, \cite{amir2018highlights} and \cite{kim2014bayesian} describe a process to construct an informative subset that represents the network's decision function. In vision, methods range from visualizing feature filters \citep{olah2017feature} to individual pixels \citep{selvaraju2016grad,selvaraju2017grad,bach2015pixel}. Similarly, in natural language processing, attention is commonly visualized to pick out informative spans for a transfer task \citep{vashishth2019attention,serrano2019attention}.
Interpretability however, has had less emphasis in contrastive learning. The closest line of research investigates the utility of contrastive representations on different transfer tasks (such as segmentation or detection) and different transfer distributions to tease out what information an embedding contains. A contribution of our work is to bring ideas from interpretability on supervised models (like informative samples) to contrastive learning. We do so by visualizing uncertainty.

\textbf{Anomaly Detection.}
We compared our method to several standard baselines, which are explained in \cite{sugiyama2013rapid}.
Most similar to our approach is a representation learning algorithm called REPEN \citep{pang2018learning}. Notably, REPEN assumes an initial classification of examples into inlier and outlier classes (in practice, it uses a distance heuristic i.e. LeSiNN to do this). REPEN then optimizes a triplet loss to learn a representation on which outliers can be further separated by distance in representation space. That is, given a deterministic encoding function $g_\theta: X \rightarrow \mathbf{R}^d$, and an inlier example $x_i$, REPEN samples a second inlier example $x_p$ and an outlier example $x_n$ using the LeSiNN classification. The REPEN loss is $\mathcal{L}_{\textup{REPEN}}(x_i) = \max(g_\theta(x_i)^T g_\theta(x_p) - g_\theta(x_i)^T g_\theta(x_n) + \alpha, 0)$
where $\alpha$ is a margin hyperparameter.
We note several similarities of REPEN to SimCLR: if we let $k=1$, $\alpha=0$, and $\tau=1$, we observe $\log \left( \frac{e^{g_\theta(x_i)^T g_\theta(x_p) / \tau}}{ e^{g_\theta(x_i)^T g_\theta(x_n) / \tau}} \right)  = g_\theta(x_i)^T g_\theta(x_p) - g_\theta(x_i)^T g_\theta(x_n)$. That is, the two are equivalent.
In practice, we find the performance of REPEN to be  contingent on LeSiNN. In comparison, we find DUM to be a much simpler and effective approach, especially in cases where LeSiNN lacks.

\textbf{Out-of-Distribution Detection.}
The vast majority of OOD algorithms derive outlier scores on top of predictions made by a large supervised neural networks trained on the inlier dataset. \cite{hendrycks2016baseline} first proposed using the maximum softmax probability from the supervised network as the outlier score. This was successively improved by ODIN \citep{liang2017enhancing}, which added temperature scaling and small perturbations via gradients to more distinctly separate inlier and outlier scores. Following this, \cite{lee2018simple} imposed Gaussian density estimates on top of the supervised network's intermediate layer activations to derive a score, which \cite{zisselman2020deep} generalized to more complex families of distributions with invertible flows. Next, \cite{sastry2019detecting} reached near ceiling performance by using Gram matrices to summarize multiple activation maps at once.
Finally, most recent is a self-supervised method proposed by \cite{hendrycks2019using} that adds a rotation prediction objective in training the supervised network.
While these methods work very well, reaching ceiling performance, they require class information, which may  be unreasonable as annotations are not always practical.
In our experiments, we find our method to have mostly comparable performance despite having zero labeled examples.

\section{Conclusion}
We introduced a simple procedure, derived from the idea of contrasting distributions, that extends existing contrastive representation learning algorithms to capture uncertainty over the learned embeddings. We explored several applications, finding that the Deep Uncertainty Model framework yields an intuitive metric of representation quality and provides strong performance on three related domain-shift detection problems: classical anomaly detection, detecting images corrupted with visual noise, and unsupervised visual out-of-distribution detection.
Future work should make theoretical connections between our method and deep generative models, explore richer uncertainty models with stronger objectives, and embedding uncertainty in new applications.


\bibliography{iclr2020_conference}
\bibliographystyle{iclr2020_conference}

\newpage
\appendix
\section{Appendix}

We provide details on experimental procedures from the main text and a few auxiliary experiments.

\subsection{Details of Visualization Experiments}
While this is stated in the main text, we emphasize the fact that we use a pretrained SimCLR model from Google's public repository: \url{https://github.com/google-research/simclr}. This characterizes one of the strengths of the DUM approach: it can leverage existing algorithms as they are. This implementation of SimCLR used slightly different data augmentations than ones we trained from scratch: it does not normalize the images and does not center crop images to 224 by 224 pixels during evaluation. Furthermore, this implementation uses a ResNet50x1 encoder. To train DUM on ImageNet representations, we optimize the DUM objective for 50 epochs using SGD with learning rate 0.01, batch size 256, momentum 0.9, and no weight decay. The input to the DUM model are the post-pooling ResNet50 features (2048 dimensions) after the final convolutional layer. The DUM encoder is a 3-Layer MLP with 4096 hidden dimensions.

\subsection{Additional Visualizations}
\label{sec:sup:viz}
\begin{figure}[h!]
    \centering
    \begin{subfigure}[b]{0.49\textwidth}
        \centering
        \includegraphics[width=\textwidth]{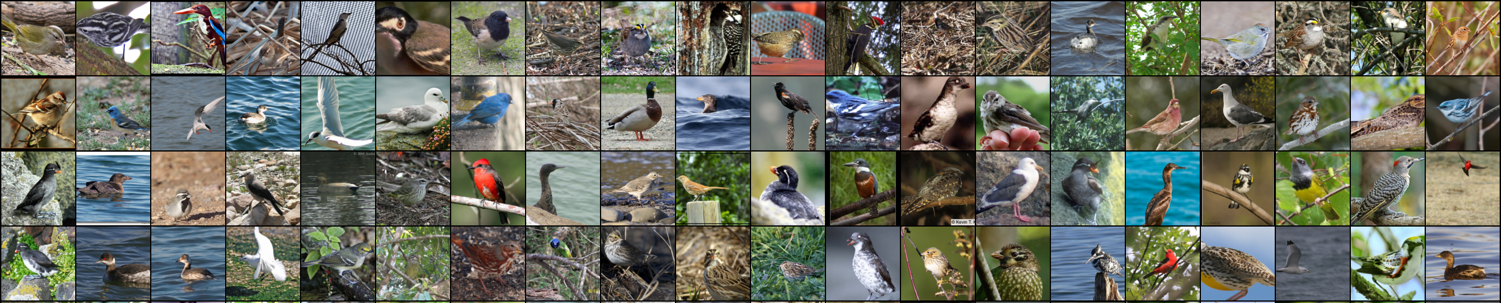}
        \caption{CUBirds: High Variance}
    \end{subfigure}
    \begin{subfigure}[b]{0.49\textwidth}
        \centering
        \includegraphics[width=\textwidth]{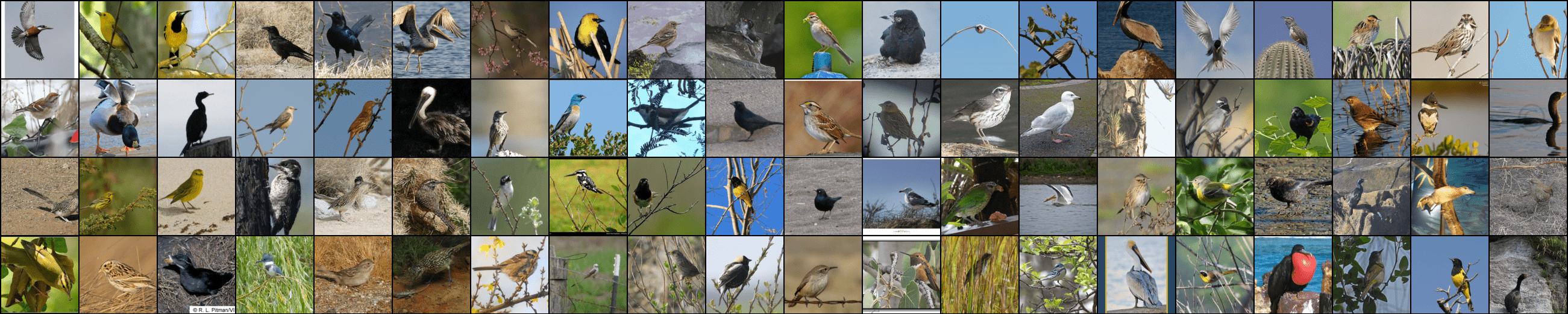}
        \caption{CUBirds: Low Variance}
    \end{subfigure}
    \begin{subfigure}[b]{0.49\textwidth}
        \centering
        \includegraphics[width=\textwidth]{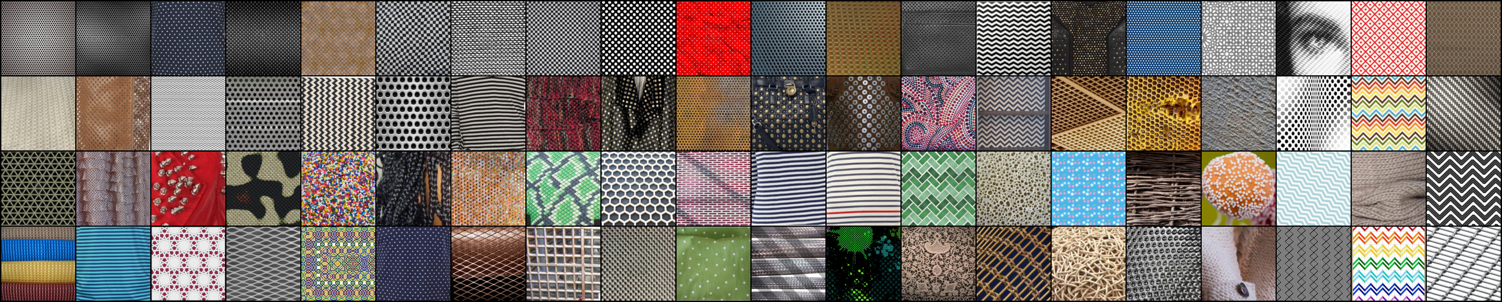}
        \caption{DTD: High Variance}
    \end{subfigure}
    \begin{subfigure}[b]{0.49\textwidth}
        \centering
        \includegraphics[width=\textwidth]{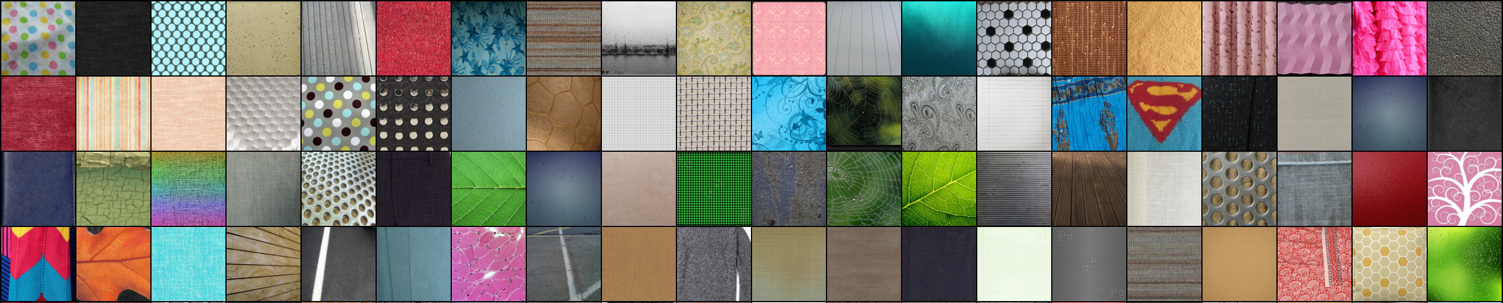}
        \caption{DTD: Low Variance}
    \end{subfigure}
    \begin{subfigure}[b]{0.49\textwidth}
        \centering
        \includegraphics[width=\textwidth]{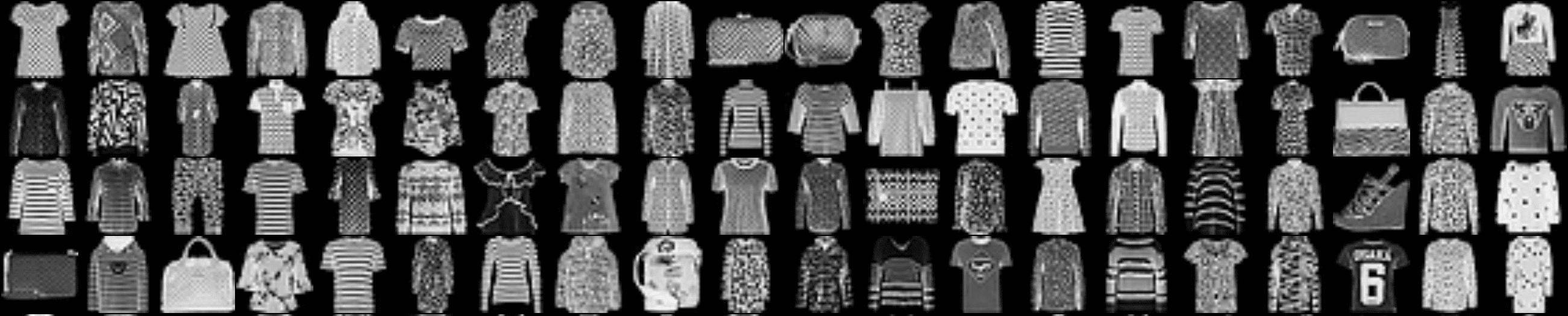}
        \caption{FashionMNIST: High Variance}
    \end{subfigure}
    \begin{subfigure}[b]{0.49\textwidth}
        \centering
        \includegraphics[width=\textwidth]{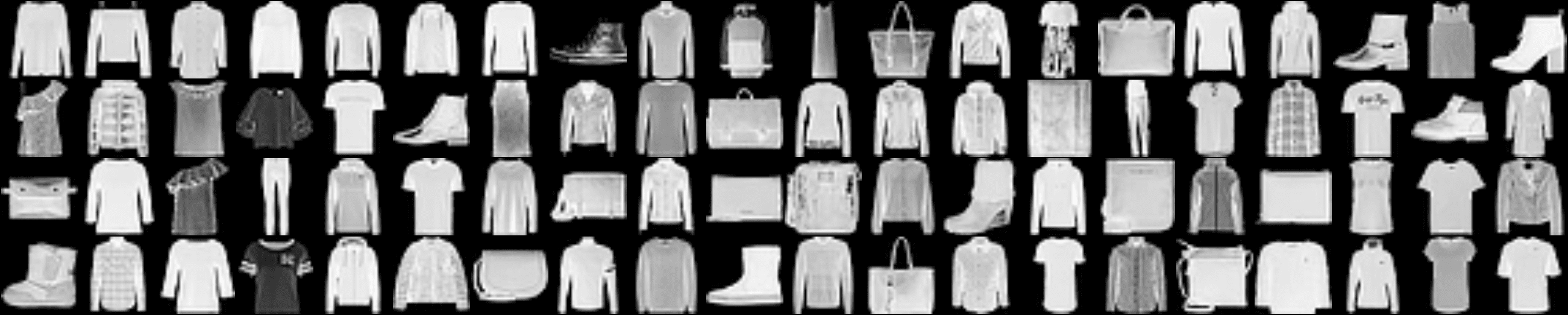}
        \caption{FashionMNIST: Low Variance}
    \end{subfigure}
    \begin{subfigure}[b]{0.49\textwidth}
        \centering
        \includegraphics[width=\textwidth]{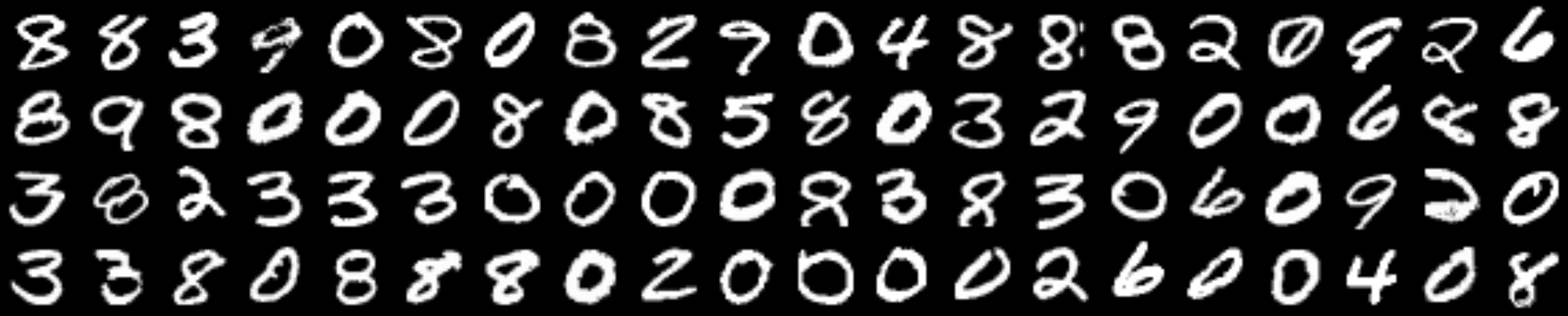}
        \caption{MNIST: High Variance}
    \end{subfigure}
    \begin{subfigure}[b]{0.49\textwidth}
        \centering
        \includegraphics[width=\textwidth]{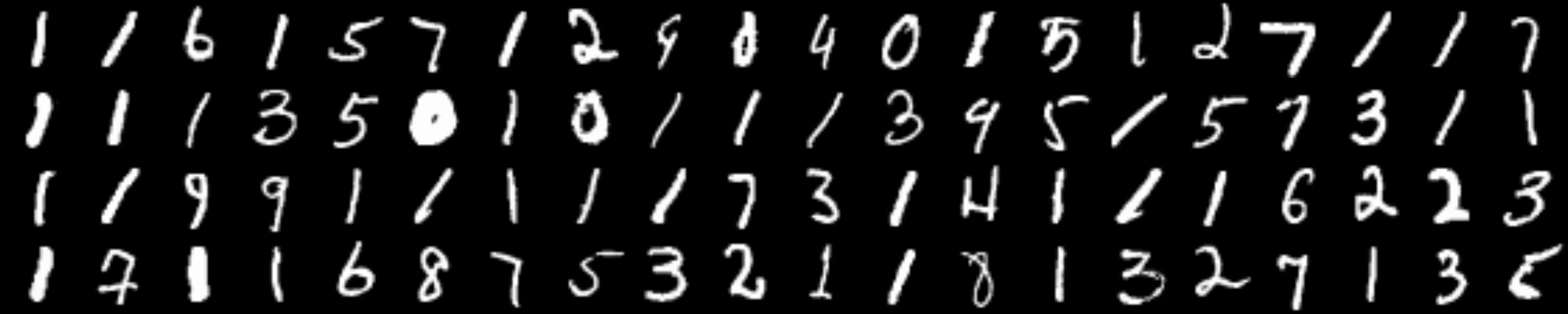}
        \caption{MNIST: Low Variance}
    \end{subfigure}
    \begin{subfigure}[b]{0.49\textwidth}
        \centering
        \includegraphics[width=\textwidth]{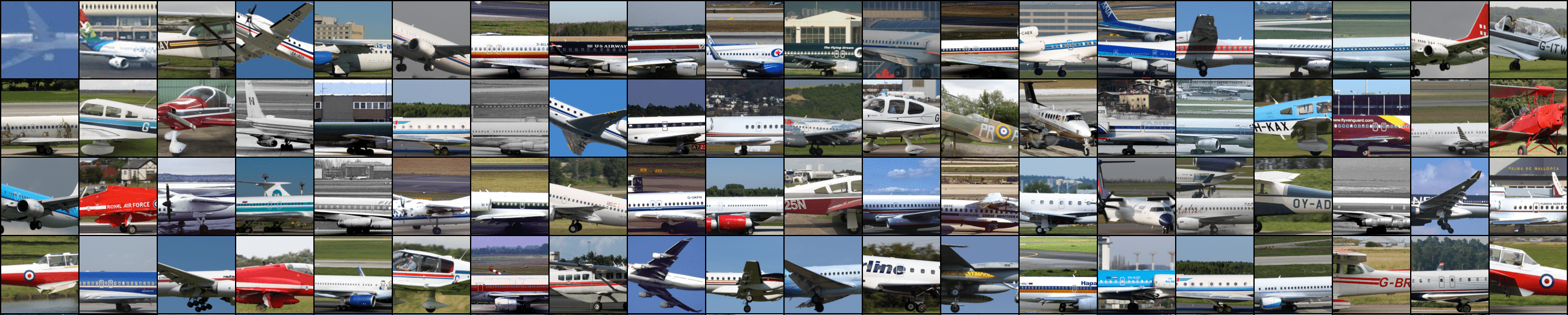}
        \caption{Aircraft: High Variance}
    \end{subfigure}
    \begin{subfigure}[b]{0.49\textwidth}
        \centering
        \includegraphics[width=\textwidth]{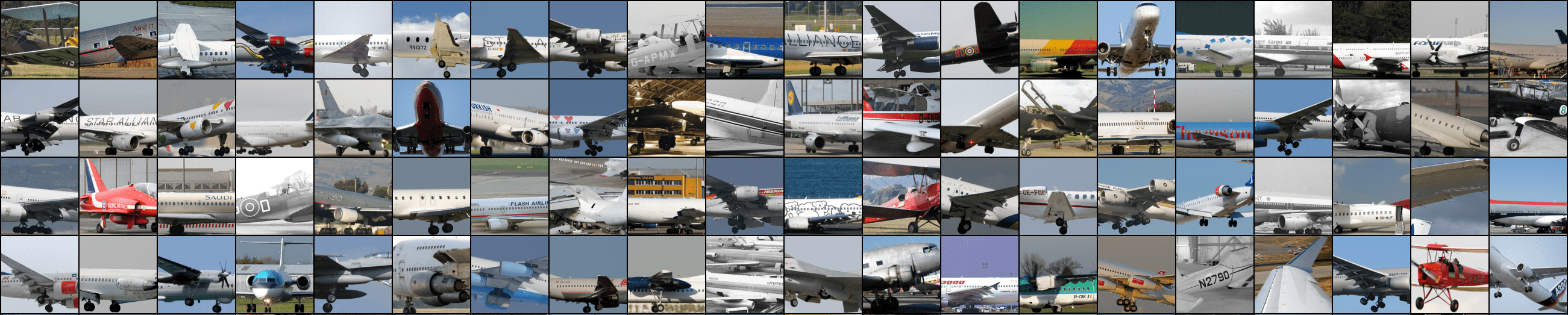}
        \caption{Aircraft: Low Variance}
    \end{subfigure}
    \begin{subfigure}[b]{0.49\textwidth}
        \centering
        \includegraphics[width=\textwidth]{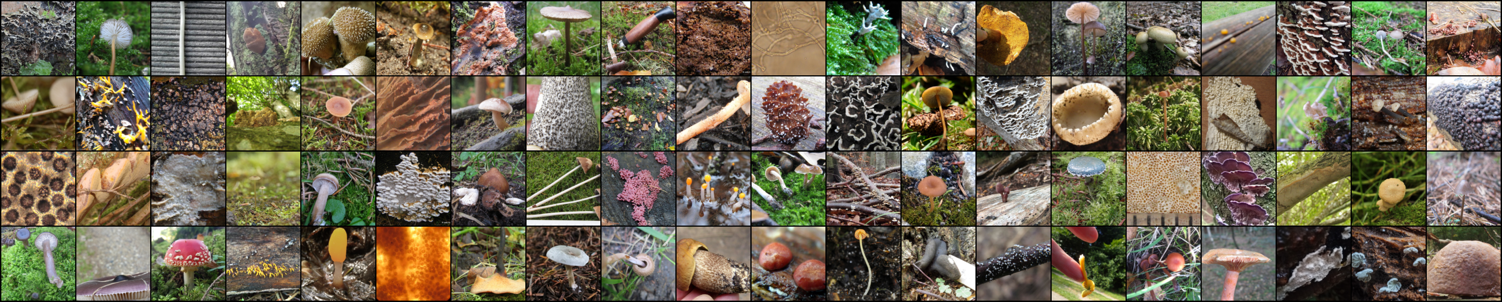}
        \caption{Fungi: High Variance}
    \end{subfigure}
    \begin{subfigure}[b]{0.49\textwidth}
        \centering
        \includegraphics[width=\textwidth]{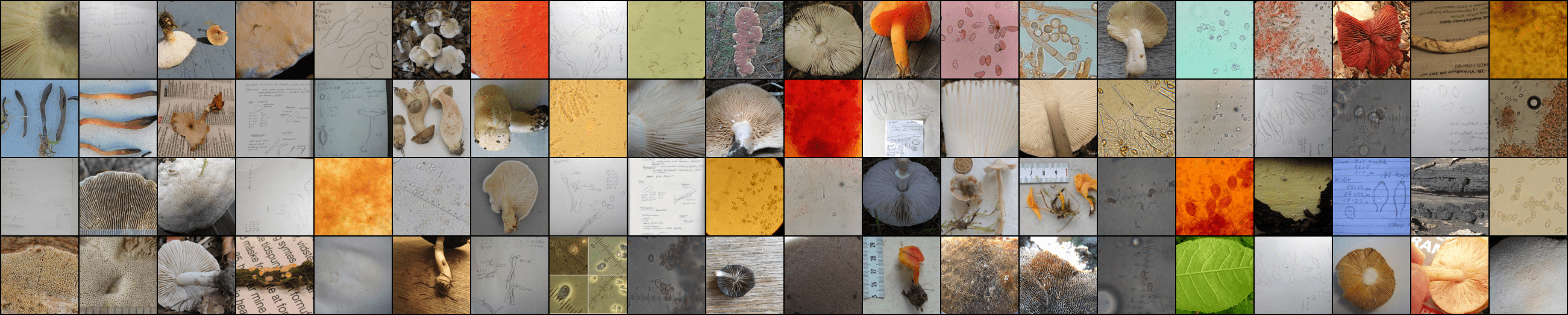}
        \caption{Fungi: Low Variance}
    \end{subfigure}
    \begin{subfigure}[b]{0.49\textwidth}
        \centering
        \includegraphics[width=\textwidth]{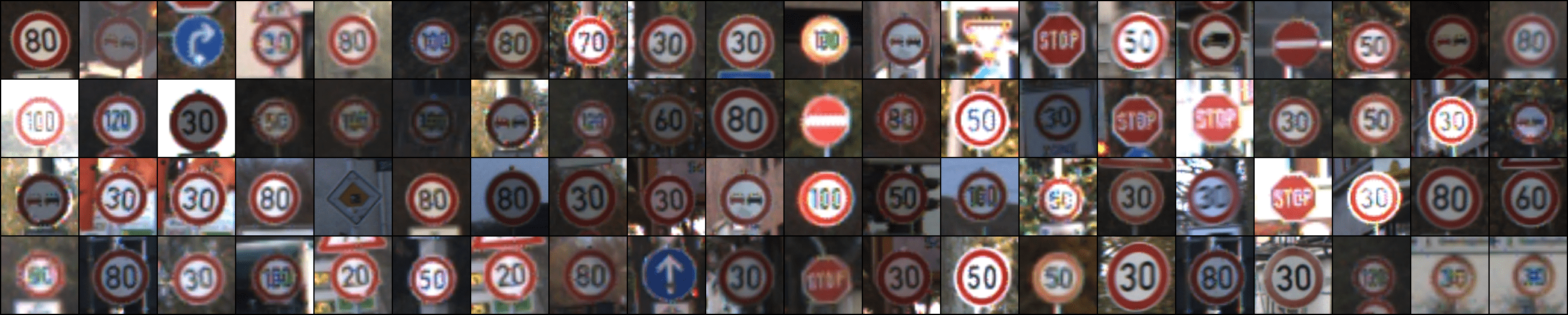}
        \caption{TrafficSign: High Variance}
    \end{subfigure}
    \begin{subfigure}[b]{0.49\textwidth}
        \centering
        \includegraphics[width=\textwidth]{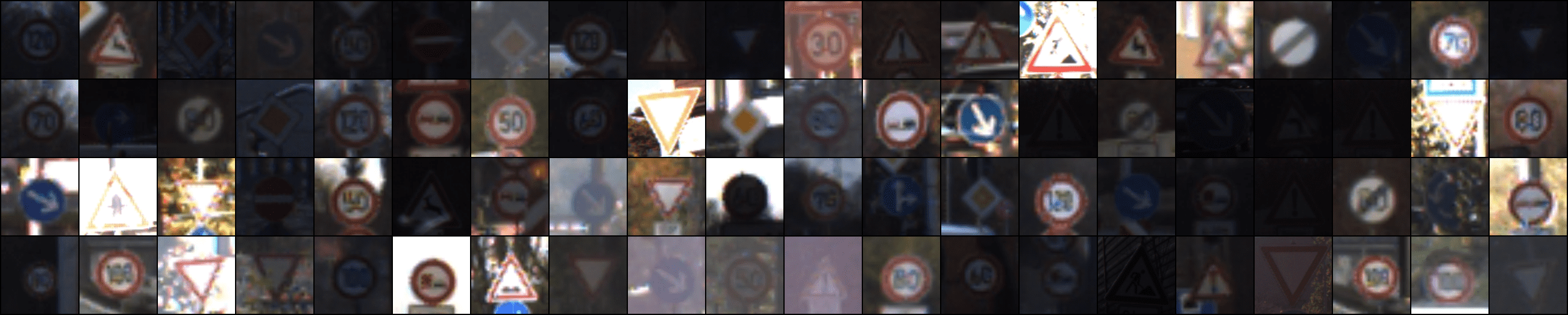}
        \caption{TrafficSign: Low Variance}
    \end{subfigure}
    \begin{subfigure}[b]{0.49\textwidth}
        \centering
        \includegraphics[width=\textwidth]{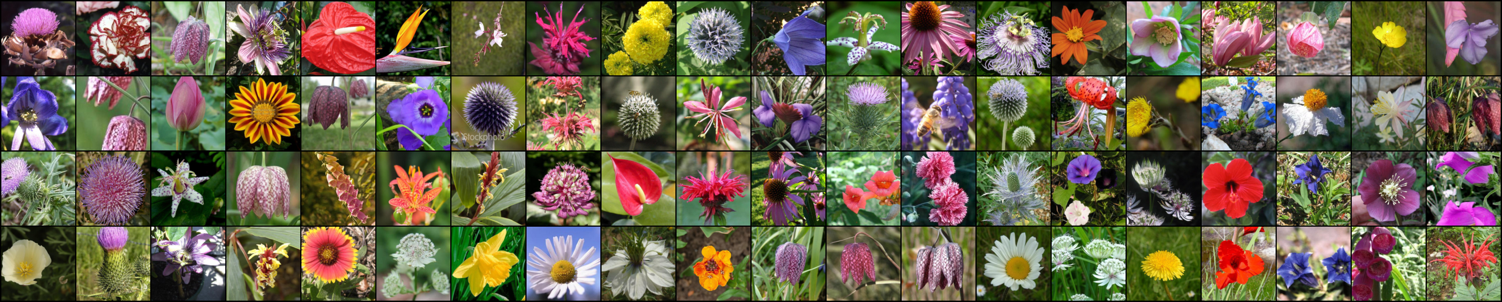}
        \caption{VGGFlower: High Variance}
    \end{subfigure}
    \begin{subfigure}[b]{0.49\textwidth}
        \centering
        \includegraphics[width=\textwidth]{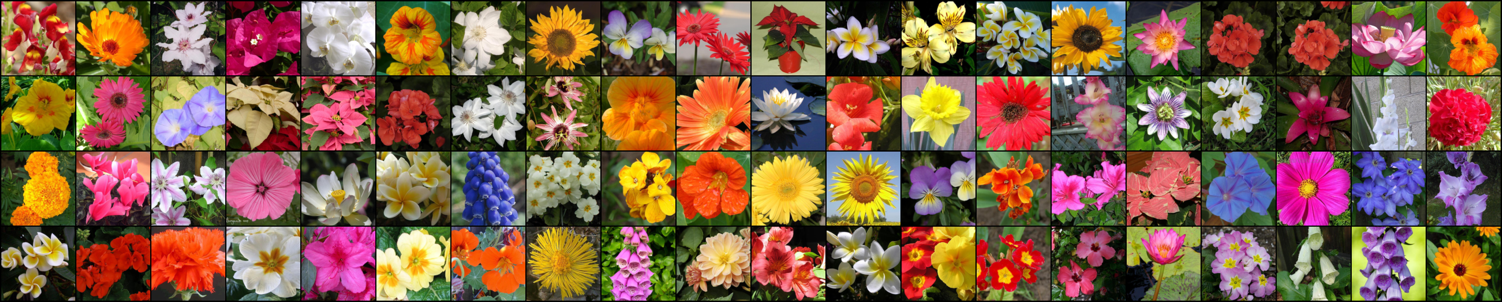}
        \caption{VGGFlower: Low Variance}
    \end{subfigure}
    \begin{subfigure}[b]{0.49\textwidth}
        \centering
        \includegraphics[width=\textwidth]{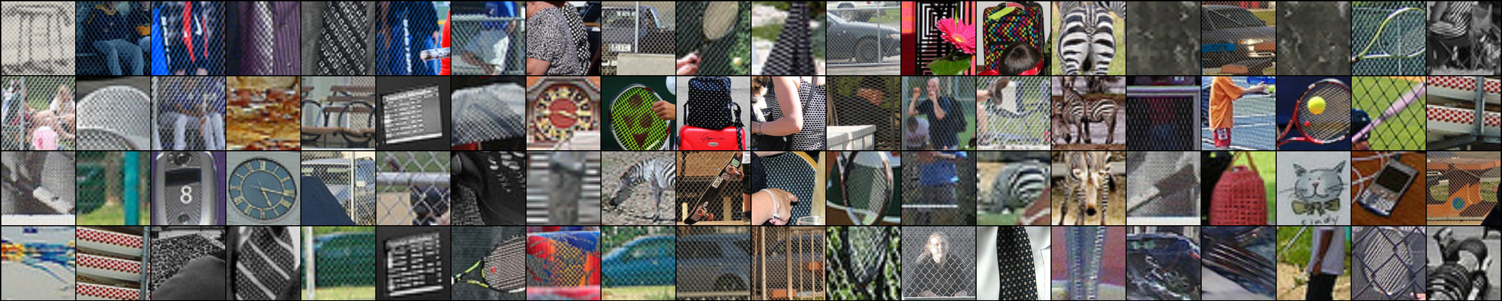}
        \caption{MSCOCO: High Variance}
    \end{subfigure}
    \begin{subfigure}[b]{0.49\textwidth}
        \centering
        \includegraphics[width=\textwidth]{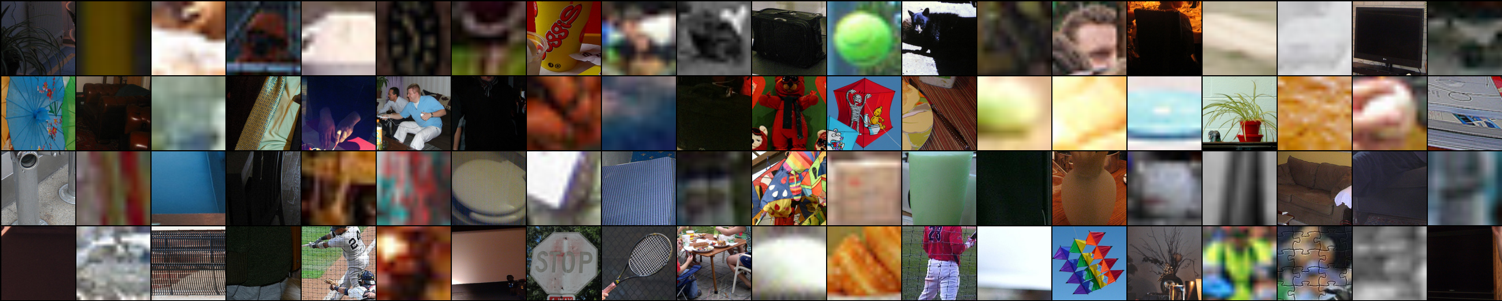}
        \caption{MSCOCO: Low Variance}
    \end{subfigure}
    \begin{subfigure}[b]{0.49\textwidth}
        \centering
        \includegraphics[width=\textwidth]{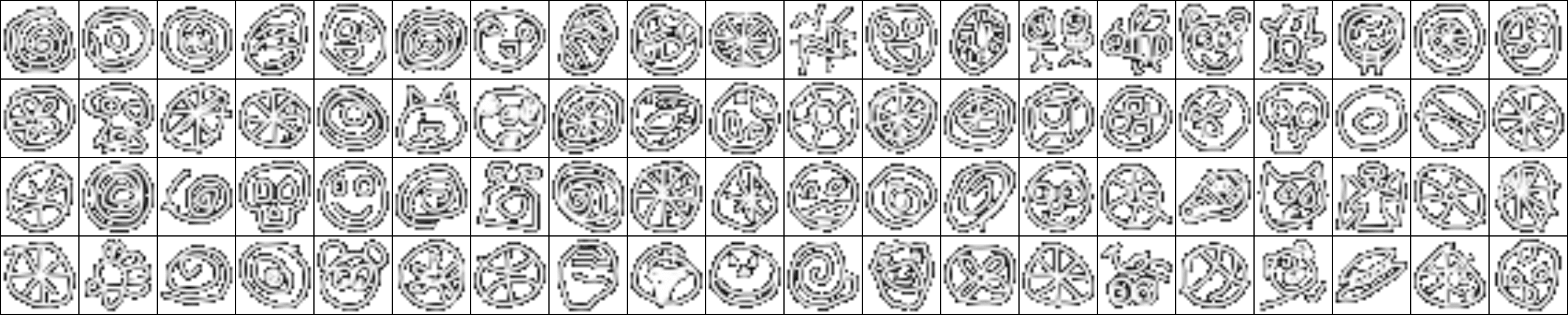}
        \caption{Quickdraw: High Variance}
    \end{subfigure}
    \begin{subfigure}[b]{0.49\textwidth}
        \centering
        \includegraphics[width=\textwidth]{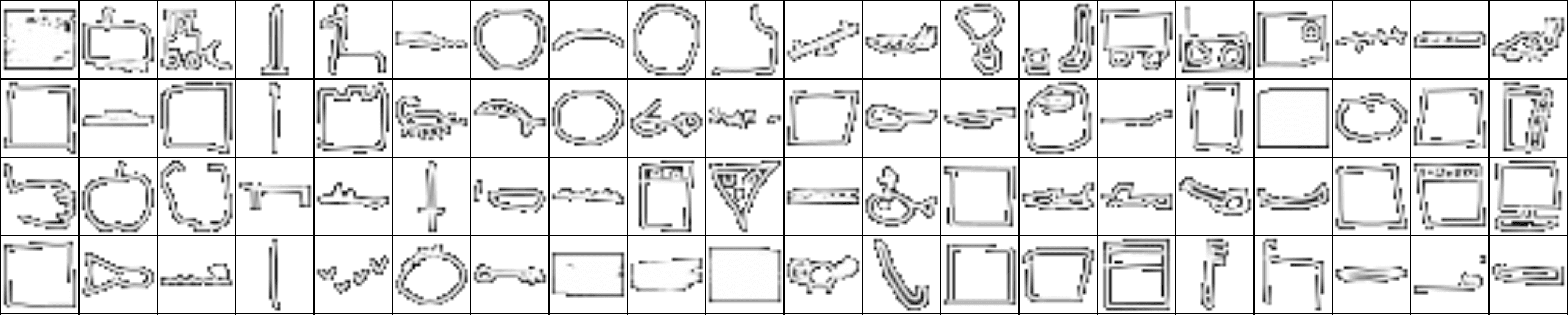}
        \caption{Quickdraw: Low Variance}
    \end{subfigure}
    \caption{Top and bottom 80 Images sorted by the norm of the variance predicted by the variational encoder for datasets in the \textit{Meta-Dataset} collection \citep{triantafillou2019meta}.}
    \label{fig:interp3}
\end{figure}

\begin{figure}[h!]
    \centering
    \includegraphics[width=\textwidth]{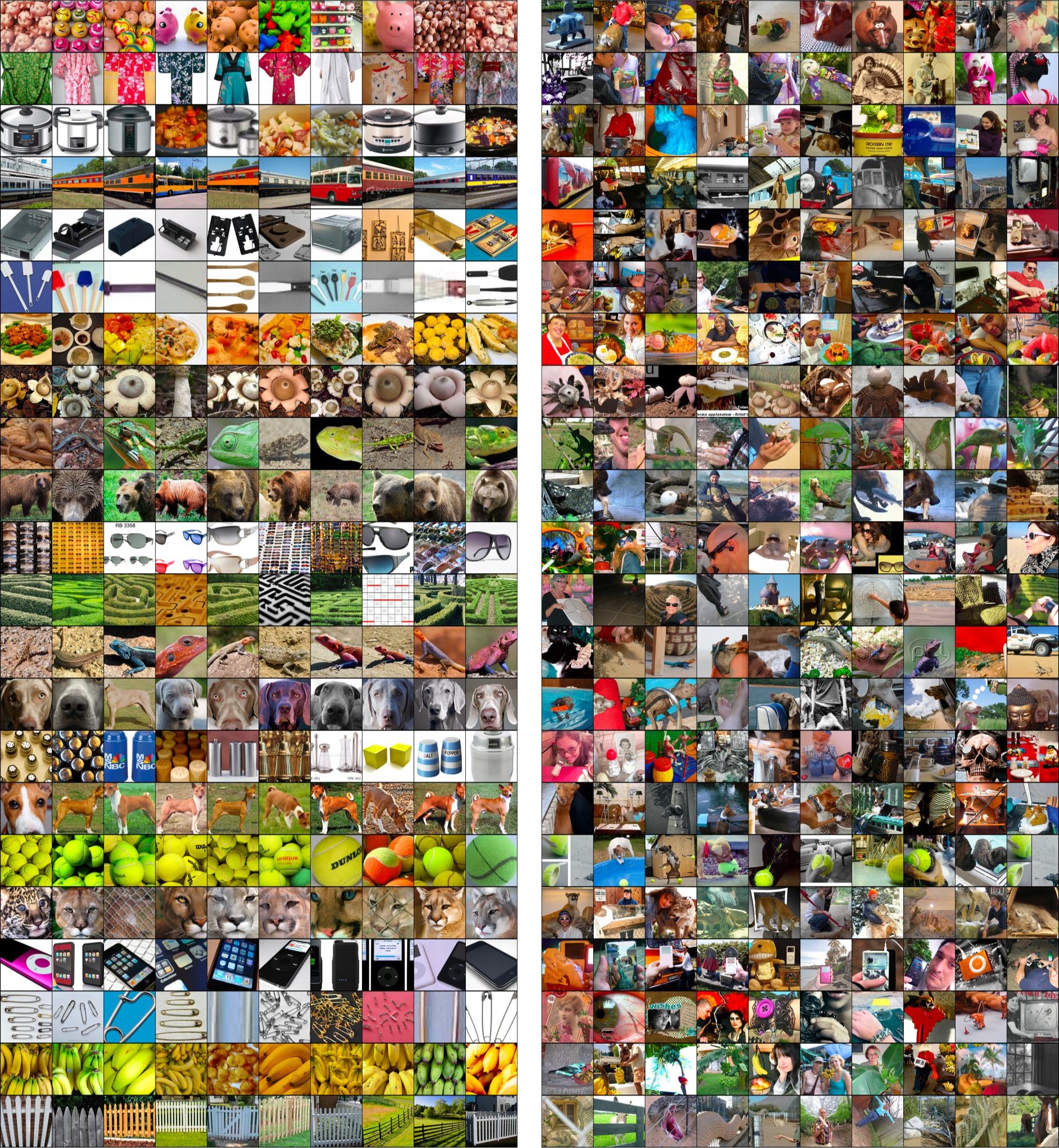}
    \caption{Expanded set of ImageNet classes showing highest and lowest DUM variance norms.}
    \label{fig:interp4}
\end{figure}

We include a more expansive set of visualizations showing the least and most certain examples to embed sorted by variance norm of the encoded distribution. Fig.~\ref{fig:interp4} show more classes chosen randomly from ImageNet whereas Fig.~\ref{fig:interp3} shows 80 of the images with the lowest and highest variance norm for 10 datasets in the Meta-Dataset collection \citep{triantafillou2019meta}. Note that these variances were extracted with a ResNet18 encoder pretrained on CIFAR10, which suggests that the features captured generalize to varied image distributions.

\subsection{Details of Corruption Experiments}
CIFAR10-C, CIFAR100-C, and TinyImageNet-C datasets were downloaded from \url{https://github.com/hendrycks/robustness}. The standard TinyImageNet dataset, which we need for its test set, was found at \url{https://tiny-imagenet.herokuapp.com}. All hyperparameters for ImageNet are as in the visualiation experiments, detailed above. For CIFAR10 and CIFAR100, we train a ResNet18 encoder using the SimCLR objective with output dimension 128. We use SGD with batch size 128, learning rate 0.03, momentum 0.9, weight decay 1e-4 for 200 epochs with no learning rate dropping. The data augmentations we use are a composition of random cropping to 224 by 224 pixels, random color jitter, random horizontal flipping, and random grayscale, plus normalization using dataset statistics. For ResNet18, we use the pre-pooling features after the last convolutional layer as the input to the DUM model. All following details are as above. To conduct the two-sample t-test, we use \textsc{scipy.stats.ttest\_ind}. To compute AUROC, we use \textsc{sklearn.metrics.roc\_auc\_score}.


\subsection{Details on Anomaly Detection Experiments}

We first describe the preprocessing for each dataset, which we found to not be obvious from prior literature. For Arrhythmia, all entries with missing data (denoted by ``?'') were replaced with 0. Everything except class 1 is considered to be an outlier. For Covertype, PIMA, SpamBase, and Skin, the least frequent class is chosen as the outlier. For Ionosphere, if the label is ``g'', it is considered in outlier. For Isolet, we use the split \textsc{isolet1+2+3+4.data} and treat classes ``C'', ``D'', and ``E'' as outliers. Note that we do not only use 10 instances of each class. All other classes are inliers. For KDD1999, we treat the \textsc{logged\_in} column as the outlier label. In addition, we ignore the following columns as they contain categorical, duplicate, or label information: \textsc{num\_outbound\_cmds}, \textsc{label}, \textsc{is\_host\_login}, \textsc{protocol\_type}, \textsc{service}, \textsc{flag}, \textsc{land}, \textsc{is\_guest\_login}. For MFeat, we concatenate the following file contents into one: \textsc{fac}, \textsc{fou}, \textsc{kar}, \textsc{mor}, \textsc{pix}, \textsc{zer}. Classes 6 and 9 are considered inliers whereas class 0 is considered an outlier (again we do not only choose 10 points of class 0). For OptDigits, we consider classes 3 and 9 as inliers and all of class 0 as outliers. For PAMAP2, we concatenate all subject files from 1 to 10 and drop the third column as it contains too much missing data. The second column is treated as the outlier label. For Record, we concatenate data in blocl files 1 to 10. We drop columns \textsc{cmp\_fname\_c2} and \textsc{cmp\_lname\_c2} and replace all remaining missing entries, denoted as ``?'' with zero. We take the least frequent class (over the last column) as the outlier label. For StatLog, all rows with missing data are discarded, following which the least frequent class is chosen as the outlier. Finally, for WDBC, the second column contain the outlier labels, where the label ``B'' is treated as an outlier and all other labels are inliers. This preprocessing procedure is largely based on the one described in \cite{sugiyama2013rapid}.

We use the implementations of ISO, LOF, SVM, EE found in scikit-learn package \cite{scikit-learn} in the following packages: \textsc{sklearn.ensemble.IsolationForest}, \textsc{sklearn.neighbors.lof.LocalOutlierFactor}, \textsc{sklearn.svm.OneClassSVM}, \textsc{sklearn.covariance.EllipticEnvelope}. We found it unfair to give the models knowledge of the contamination rate, which is unknown in real world contexts. For KNN, ABOD, and AE, we use the implementations found in the Python toolkit for detecting outlying objects, PyOD \citep{zhao2019pyod}. For the autoencoder, we use a batch size of 32 if the datasize is less than 10k entries, otherwise a batch sie of 256. In the first case, we trainf or 100 epochs whereas we train for 20 in the latter. The architecture of the AE is an MLP with the hidden sizes 16, 8, 8, 16. We base our PyTorch implementation of RAMODO/REPEN after the public implementation found at \url{https://github.com/GuansongPang/deep-outlier-detection}, although with significant refactoring. We use Adam optimizer with a learning rate of 1e-3, weight decay of 1e-5, batch size 256 and 30 epochs. For our proposed method, we use an MLP with three layers, each with 4096 hidden nodes and followed by ReLU nonlinearity. We optimize with Adam with a learning rate of 1e-3, batch size 256, and a temperature of 0.07. For REPEN and our method, we train for 5 epochs only for very large datasets like PAMAP2 or KDD1999. In RAMODO, we intialize the elements in the outlier set with a subsample size of 8 and an ensemble size of 50. The KDTree uses a euclidean metric.

\subsection{Additional Experiments for Anomaly Detection}
\label{sec:app:simclr}
We mentioned in the main text that training DUM on top of the raw features performs about the same as DUM on top of SimCLR embeddings train on the raw features. Table~\ref{table:anomaly2} shows results using SimCLR embeddings for a subset of the 14 datasets below (chosen for speed).
\begin{table}[h!]
\centering
\tiny
\caption{Lesion: comparing DUM with and without SimCLR features.}
\begin{tabular}{l|l|l|c|c}
\toprule
\multicolumn{5}{c}{Area under the Receiver Operating Characteristic (AUROC)} \\
\midrule
& \# data & \# out & DUM+SimCLR & DUM \\
\midrule
Arrhythmia & 452 & 207 & $76.0$ & $76.6$ \\
Ionosphere & 351 & 126 &  $82.1$ & $81.0$ \\
Isolet & 960 & 240 & $99.9$ & $100.0$ \\
MFeat & 600 & 200 & $99.1$ & $99.9$ \\
OptDigits & 1.7K & 554 & $95.4$ & $99.6$ \\
PIMA & 768 & 268 & $82.0$ & $81.5$ \\
Spambase & 4.6K & 1.8K & $82.5$ & $83.4$ \\
Statlog & 6.4K & 626 & $84.7$ & $89.2$ \\
Wdbc & 569 & 212 & $96.2$ & $96.9$ \\
\bottomrule
\end{tabular}
\label{table:anomaly2}
\end{table}

\subsection{Details on Out-of-Distribution Detection Experiments}
We first train SimCLR on each of the inlier distributions using a ResNet34 encoder (to be comparable to supervised baselines, which all use ResNet34), temperature $\tau=0.07$, and an embedding dimension of 128. For optimization we use SGD, momentum 0.9, learning rate 0.03, batch size 128 for 200 epochs. All images are resized to 256 by 256 prior to augmentations. After this, we fit DUM on learned embeddings, using the same MLP architecture and hyperparameters as in Sec.~\ref{section:anomaly}.
Our implementation of baselines is heavily based on the following public github repositories: \url{https://github.com/pokaxpoka/deep\_Mahalanobis\_detector}, \url{https://github.com/EvZissel/Residual-Flow}, and \url{https://github.com/hendrycks/ss-ood}, \url{https://github.com/VectorInstitute/gram-ood-detection}, which in total contain implementations for all six baselines. In addition, these baselines contain pretrained backbone networks on CIFAR10, CIFAR100, and SVHN, which we download and utilize in our replications of their results. The LSUN and TinyImageNet dataset splits were downloaded from the Mahalanobis public repository. For Rotation Prediction, we pretrain the joint supervised and contrastive objective with 0.5 weight on the rotation objective and 0.5 weight on the translation objective. We use SGD with a learning rate of 0.1, momentum 0.9, weight decay 0.0005, batch size 32 for 50 epochs with linear learning rate scheduling. For our proposed method, we optimize SimCLR with ResNet32 for 200 epochs using SGD, momentum 0.9, weight decay 1e-4. The representation dimensionality is 128, and we use a temperature of 0.07. Following this, we train DUM for 100 epochs, using Adam with learning rate 1e-3.

\end{document}